\documentclass[runningheads]{llncs}
\usepackage[T1]{fontenc}
\usepackage{booktabs}
\usepackage{multicol}
\usepackage{multirow}
\usepackage{graphicx}
\usepackage{caption}
\usepackage{subcaption}
\usepackage{algorithm}
\usepackage{xcolor}
\usepackage[noend]{algpseudocode}
\usepackage{color-edits}
\addauthor{cm}{blue}
\addauthor{sw}{orange}

\begin{document}
%
\title{Generating Situated Reflection Triggers about Alternative Solution Paths: A Case Study of Generative AI for Computer-Supported Collaborative Learning}
%
%
\author{Atharva Naik \and Jessica Ruhan Yin \and Anusha Kamath \and Qianou Ma \and Sherry Tongshuang Wu \and Charles Murray \and Christopher Bogart \and Majd Sakr \and Carolyn P. Rose}

\authorrunning{F. Author et al.}

%
\institute{Carnegie Mellon University, Pittsburgh PA 15213, USA \\
\url{https://www.cmu.edu/}
}
\maketitle              
\begin{abstract}
An advantage of Large Language Models (LLMs) is their \emph{contextualization} capability -- providing different responses based on student inputs like solution strategy or prior discussion, to potentially better engage students than standard feedback.
We present a design and evaluation of a proof-of-concept LLM application to offer students dynamic and contextualized feedback.
Specifically, we augment an Online Programming Exercise bot for a college-level Cloud Computing course with ChatGPT, which offers students contextualized reflection triggers during a collaborative query optimization task in database design. 
We demonstrate that LLMs can be used to generate highly situated reflection triggers that incorporate details of the collaborative discussion happening in context.  
We discuss in depth the exploration of the design space of the triggers and their correspondence with the learning objectives as well as the impact on student learning in a pilot study with 34 students. 

\end{abstract}
\section{Introduction}
For nearly two decades intelligent conversational agents have been employed to increase reflection and learning in Computer-Supported Collaborative Learning (CSCL) settings \cite{gweon_cscl_feedback,kumar2007tutorial,tegos2015promoting,rose2016technology,sankaranarayanan2022collaborative}.  
This line of work has yielded principles for the design of interactive collaborative scaffolding that increases learning impact over static forms of scaffolding \cite{vogel_static_scaffolding}, which is prevalent in the field of CSCL.  
In particular, the positive impact of situating reflection triggers in a specific conversational context has been established in earlier studies \cite{ai2010finding,cui2009helping}.   
Recent studies \cite{sree_reflection_better} focusing specifically on Computer Science (CS) education suggest that collaboration support that shifts the focus of students more toward reflection and less towards the actual coding increases conceptual learning without harming the ability to write code in subsequent programming assignments.  
These past studies focused on manipulating the timing of the reflection triggers or the proportion of time dedicated to reflection versus programming.  
Recent advances in Generative AI (GenAI) and Large Language Models (LLMs) have enhanced AI capabilities for the evaluation of multimodal student input and real-time feedback, which has provoked intensive exploration of the space of application possibilities \cite{kasneci_seßler_küchemann_bannert_dementieva_fischer_gasser_groh_günnemann_hüllermeier_et}. 
This technology opens up more options for adapting the specific content of reflection triggers from specific details of the students' work and discussion in context.

So far, less attention has been given to applying principles for designing effective dynamic support for CSCL using advances in GenAI. Most explorations are geared towards individual programmers, such as programming assistance for individual novice learners \cite{LearnabilityofLLMs,AICodeGenForNovice,gen_prog_trivially,UsabilityLLMCode,kazemitabaar2023novices}. 
However, past success in developing intelligent forms of effective scaffolding for collaborative learning suggests that advances leading to the prevalence of these LLMs of code create a ripe area for exploration.  The focus of this paper is the application of GenAI to dynamic support for reflection in learning through collaborative software development. 

Our key contributions are outlined below:
\begin{itemize}
    \item \emph{Technical contribution:} Development of a novel prompt engineering approach to elicit contextually appropriate suggestions of alternative code contributions from code LLMs as a dynamic form of reflection trigger.
    \item \emph{Learning resource development contribution:} Enhancement of a platform for online collaborative software development with dynamic support for reflection during software development.
    \item \emph{Learning research contribution:} Testing the impact of LLM-constructed reflection triggers on student learning in an online collaborative SQL optimization activity, along with a thorough analysis of their impact on learning.
\end{itemize}
\section{Related Work}

\subsection{Generative AI in Education}
Since its December 2022 launch, ChatGPT has emerged as a leading GenAI technology \cite{Duarte,Carr}. 
Its accessibility has sparked innovations and debates on its role in education \cite{ChatGPTStudentDriven}. 
For students, its ability to process cross-domain knowledge is appealing \cite{ChatGPT_GenAI_Science}, while educators see potential in content creation \cite{young2023evaluation}, personalized tutoring \cite{sridhar2023harnessing}, as well as risks of enabling plagiarism \cite{ChatGPTPlagiarism1} and dissemination of biased information \cite{sok2023chatgpt}.
Mixed sentiments surround GenAI's impact on academic integrity and the future of education \cite{perceptions_genAI}.
Despite the potential risks, some education technologists are more optimistic and view GenAI as a student-centric technology for personalization and customized real-time feedback \cite{ChatGPTStudentDriven}. In our work, we provide personalized real-time intervention for collaborative learning, in the form of reflection triggers using ChatGPT.

\subsection{Intelligent Support for Collaboration}
Providing technological support for collaborative and discussion-based learning has long been the focus of CSCL \cite{rose2016technology}.
Past studies highlight the benefits of interactive and context-sensitive support in group learning \cite{kumar2007tutorial,kumar2010architecture}.
While static scaffolding like fixed prompts \cite{vogel_static_scaffolding} and scripted roles \cite{Fischer2013-xh} have been effective, contextualized interventions within specific conversational contexts \cite{ai2010finding,cui2009helping} or perceived roles of students \cite{gweon_cscl_feedback} have also shown positive outcomes.
Studies like \cite{kumar2007tutorial,kumar2010architecture,rose2016technology} have shown the effectiveness of discussion-based learning and conversational support using dialog agents.
Finally \cite{sree_reflection_better,sankaranarayanan2022collaborative} have shown the effectiveness of reflection-based learning for programming, showing that shifting the focus of students more towards reflection than actual coding can increase conceptual learning without harming the ability to write code \cite{sree_reflection_better}.
However, these studies have not explored contextualized scaffolding based on multimodal input (like code entered by students), which is now possible with LLMs like ChatGPT and is the main focus of this work.

\subsection{Intelligent Support for Programming in CS Education}
In the realm of collaborative programming, most studies involve humans collaborating with intelligent agents in pair programming paradigms \cite{ai_sub_pair_programming,kuttal_conv_agents_pair_prog}. 
Some newer approaches take a more relaxed definition of pair programming as a collaborative setting where a human programmer receives AI assistance, often dubbed as pAIr programming \cite{wu2023ai}.
Studies like \cite{UsabilityLLMCode,LearnabilityofLLMs} focus on program synthesizers, providing design recommendations to support novice programmers. 
Kazemitabaar et. al. \cite{AICodeGenForNovice} analyzed the impact the Codex LLM assistance on 69 novice learners and found improvements in code authoring performance. 
Yilmaz and Karaoglan Yilmaz \cite{YILMAZ2023100147} found a significant and positive impact of ChatGPT programming assistance on the computational thinking, programming self-efficacy, and motivation of undergraduate students.
However, in all of these studies, a sole human user pairs up with an AI agent, effectively making them individual learning scenarios. 
Our work tackles the more under-explored paradigm of facilitating collaborative learning in \textit{mob programming}, where 3 to 5 students work on the same task while playing different roles \cite{WikiMobProgDef}, which is used more often in CSCL for advanced CS topics \cite{sree_reflection_better,sankaranarayanan2022collaborative}. 
\section{Learning Activity Design}
Collaborative learning is most valuable for learning activities where there are multiple possible solution paths, and selecting from among them is less about finding the right answer than it is about evaluating complex sets of constraints and trade-offs \cite{cress2021foundations,koschmann2017computer}. Amid these activities, students benefit from exposure to each other's alternative points of view. Because CS is an engineering discipline, it is ripe with opportunities for evaluation of design trade-offs, especially in connection with advanced topics.  We select a topic with important design trade-offs and a paradigm for orchestrating the collaboration that encourages sharing and challenging alternative perspectives.

\subsection{Learning Task: SQL Optimization}
Especially when large amounts of data are involved, database design offers a solution space with interesting trade-offs. We situate our investigation in the SQL optimization task, which involves organizing and modifying the database using techniques like datatype modification, index creation, and table joining (denormalization) for a given scenario or query load, to minimize query cost while satisfying a few constraints.
The optimization centers around the following rubric dimensions: 
\\
\textbf{Data Retrieval Efficiency:} how quickly and efficiently the database can retrieve data and execute queries, and how optimization techniques/design can improve performance. 
It can benefit from techniques like denormalization and indexing in certain cases.
\\
\textbf{Write Performance:}
how effectively the database can handle insert, update, and delete operations, and how
optimization techniques/design can affect performance.
It can be hurt from indexing and denormalization.
\\
\textbf{Disk Storage:} 
how efficiently the database uses disk storage, and how optimization techniques/design can reduce storage usage and improve performance.
\\
\textbf{Maintainability:} 
how effectively the database design and optimization techniques enable the database to be maintained and updated over time, and how optimization techniques can simplify maintenance and prevent additional processing or complexity for developers.
\\
\subsection{Knowledge Resources: Primers and Learning Objectives}
The specific target learning objectives are enumerated in Table~\ref{tab:learning_objectives_and_primers}. We provide students with three different primers to prepare them for the learning activity, which is the application of the primers in a concrete task. 
\\
\textbf{Denormalization \& Normalization:} This primer covers concepts of normalization and denormalization, their benefits, and trade-offs. 
It also compares the read \& write costs for both strategies along the rubric dimensions.
\\
\textbf{Data Types in MySQL:} This primer covers MySQL data types and their trade-offs for comparisons like \texttt{CHAR vs VARCHAR}, \texttt{INT vs CHAR} and contexts where each type is helpful. 
\\
\textbf{Indexing:} This primer covers the trade-offs of indexing in databases as well as how they affect the performance of read and write operations. 
It also compares single-column and composite indexing strategies along the rubric dimensions.

\begin{table}[]
\caption{The mapping of learning objectives (LOs) and primers.}
\begin{tabular}{@{}ll@{}}
\toprule
\textbf{Learning Objective (LO)}                                                                                                                    & \textbf{Primer} \\ \midrule
L1: Comparing benefits of a single column or a composite index                                                                                      & Indexing        \\
L2: Discussing trade-offs of creating an index                                                                                                      & Indexing        \\
\begin{tabular}[c]{@{}l@{}}L3: Discussing trade-offs of using single-column and multi-column \\ composite indexes in MySQL across various use cases.\end{tabular} &
  Indexing \\
\begin{tabular}[c]{@{}l@{}}L4: Evaluating and comparing the complexity of updating data \\ in normalized and denormalized tables.\end{tabular}      & Denormalization \\
\begin{tabular}[c]{@{}l@{}}L5: Identifying use cases where denormalized or normalized \\ tables would be preferred.\end{tabular}                    & Denormalization \\
\begin{tabular}[c]{@{}l@{}}L6: Comparing the performance of queries when using \\ normalized vs. denormalized tables.\end{tabular}                  & Denormalization \\
\begin{tabular}[c]{@{}l@{}}L7: Evaluating different data types and determining the most \\ appropriate choice for a given table field.\end{tabular} & Data type       \\ \bottomrule
\end{tabular}
\label{tab:learning_objectives_and_primers}
\end{table}

\subsection{Collaborative Learning Paradigm: Mob Programming}
To intensify the opportunity for reflection, we bring additional perspectives into the discussion by adopting Mob Programming \cite{WikiMobProgDef}, where 3 to 5 students work synchronously in different roles: \textit{navigator}, \textit{driver} and \textit{researcher}.
The \textit{driver} controls the keyboard and mouse, the \textit{navigator} chooses what the driver works on, and the \textit{researcher} finds related knowledge requested by the driver.

In our deployment, the activity includes three tasks, and role assignment occurs at the onset of each such that the students cycle through all three roles.
All the tasks involve aspects of all three optimization strategies but task1, task2, and task3 have a greater focus on datatype conversion, indexing (specifically composite indexing), and denormalization (and trade-offs related to rates of reads and writes of joined tables) respectively.
The activity is orchestrated by an Online Programming Exercise bot or \texttt{OPE\_Bot}, which does several housekeeping tasks like role assignment, hints, and asking reflection questions.
\section{Method}
\begin{figure*}[!tbh]
    \centering
    \includegraphics[width=\textwidth]{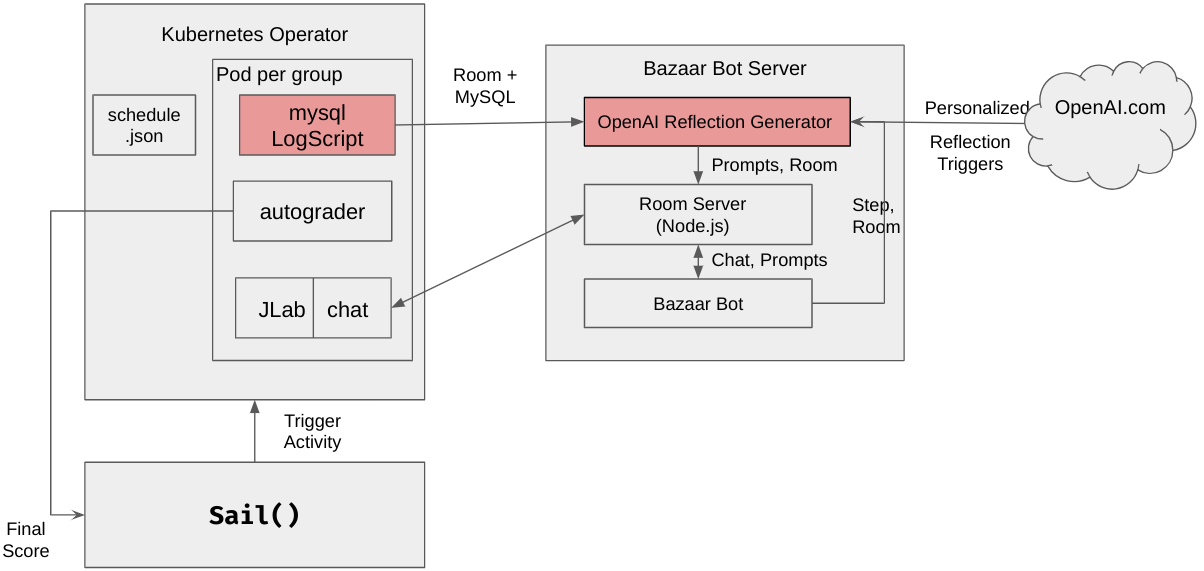}
    \caption{The existing cloud infrastructure for the Online Programming Exercise (OPE) along with our newly added components (highlighted in red) for generating dynamic and personalized reflection triggers with ChatGPT}
    \label{fig:supporting_infrastructure}
    \vspace{-10pt}
\end{figure*}

We integrate our dynamic reflection triggers into a pre-existing cloud-based system shown in figure~\ref{fig:supporting_infrastructure}.
The activity session for a group is triggered according to the schedule plan (schedule.json) by the \texttt{Sail()} platform \cite{SAIL}.
The students run their MySQL commands on a Kubernetes\footnote{\url{https://kubernetes.io/}} virtual machine via a modified JupyterLab\footnote{\url{https://jupyterlab.readthedocs.io/en/stable/}} (JLab) front-end, containing a chat window on the side for each group.
We implement a mySQL LogScript to send the necessary context like the SQL commands entered by the students to the OpenAI Reflection Generator which determines when to intervene and provide personalized reflections by prompting ChatGPT with the appropriate context.
We coordinate the activity for each group with an instance of the \texttt{OPE\_bot} which is based on the Bazaar CSCL architecture\cite{BasilicaBot}. 
When the students submit their solution, the assignment auto-grader computes a final score and sends it to the \texttt{Sail()} platform.

To realize the OpenAI Reflection Generator, we design components that determine when to intervene (reflection triggering), how to use the context like SQL commands to personalize the reflections (reflection personalization), validate the correctness of the reflections (reflection validation), and decide when to show them and how to space them apart (reflection scheduling). Each of these components is discussed in detail in sections \ref{sec:reflection_triggering}, \ref{sec:reflection_personalization}, \ref{sec:reflection_validation}, and \ref{sec:reflection_scheduling} respectively.



\begin{figure*}[!tbh]
    \centering
    \includegraphics[width=\textwidth]{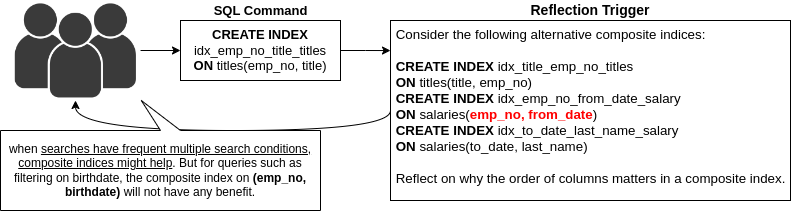}
    \caption{An example of the \texttt{COMPOSITE\_IND\_COL\_ORDER} reflection trigger along with a student response, demonstrating that our interventions make students think about the tradeoffs involved (underlined text) in the optimization.}
    \label{fig:working_example}
    \vspace{-10pt}
\end{figure*}

\subsection{Reflection Design and Triggering}
\label{sec:reflection_triggering}
To target the learning objectives from each of the primers we design five types of reflection triggers, which are triggered by specific command patterns based on regular expression-based matching as shown in table~\ref{tab:prompt_triggering}.

For the data type primer, we designed the \texttt{DATATYPE\_COMPARISON} reflection to be shown whenever students change the datatype of a column to facilitate the discussion of their choice.
For the indexing primer, we designed two reflection triggers:
the \texttt{COMPOSITE\_VS\_MULTI\_SINGLE} reflection to encourage a comparison between composite and single column indices whenever the students create a single column index and the \texttt{COMPOSITE\_IND\_COL\_ORDER} to facilitate discussion around the optimal ordering of columns whenever the students create a composite index. For the denormalization primer, the \texttt{DENORMALIZATION\_WHEN} reflection encourages a discussion about the potential costs and benefits of denormalization for queries with inner joins whenever students use such queries and the \texttt{TABLE\_CHOICE\_DENORMALIZATION} reflection encourages a discussion about the choice of tables to be joined based on the expected reads and writes for each table, whenever the students create a denormalized table.

\begin{table*}[]
\vspace{-10pt}
\caption{Five types of reflection triggers, the primers and learning objectives they address, and the student activities that trigger each type of reflection.}
\centering
\resizebox{\textwidth}{!}{%
\begin{tabular}{@{}llll@{}}
\toprule
Reflection Type &
  Student Activity Trigger &
  Primer &
  LO \\ \midrule
\texttt{DATATYPE\_COMPARISON} &
  \texttt{ALTER TABLE} \textless{}tab\textgreater \texttt{MODIFY} \textless{}col\textgreater \textless{}datatype\textgreater{} &
  Data type &
  L7 \\
\texttt{COMPOSITE\_VS\_MULTI\_SINGLE} &
  \texttt{CREATE INDEX} \textless{}ind\textgreater \texttt{ON} \textless{}tab\textgreater{}(\textless{}col\textgreater{}) &
  Indexing &
  L1, L3 \\
\texttt{COMPOSITE\_IND\_COL\_ORDER} &
 \texttt{CREATE INDEX} \textless{}ind\textgreater ON \textless{}tab\textgreater{}(\textless{}col\textsubscript{1}\textgreater{}, ... \textless{}col\textsubscript{n}\textgreater{}) &
  Indexing &
  L2 \\
\texttt{DENORMALIZATION\_WHEN} &
  \begin{tabular}[c]{@{}l@{}}\texttt{SELECT} \textless{}col\textsubscript{1}\textgreater{}, ... \textless{}col\textsubscript{n}\textgreater \texttt{FROM} \textless{}tab\textsubscript{1}\textgreater \texttt{INNER} \\ \texttt{JOIN}\textsubscript{1} \textless{}tab\textsubscript{2}\textgreater \texttt{ON} \textless{}condition\textsubscript{1}\textgreater \dots \texttt{INNER JOIN}\textsubscript{n} \\ \textless{}tab\textsubscript{n+1}\textgreater \texttt{ON} \textless{}condition\textsubscript{n}\textgreater{}\end{tabular} &
  Denormalization &
  L5, L6 \\
\texttt{TABLE\_CHOICE}\texttt{\_DENORMALIZATION} &
  \begin{tabular}[c]{@{}l@{}}\texttt{CREATE TABLE} \textless{}tab\textgreater \texttt{AS SELECT} \textless{}col\textsubscript{1}\textgreater{}, \dots \\ \textless{}col\textsubscript{n}\textgreater \texttt{FROM} \textless{}tab\textsubscript{1}\textgreater \texttt{INNER JOIN}\textsubscript{1} \textless{}tab\textsubscript{2}\textgreater \texttt{ON} \\ \textless{}condition\textsubscript{1}\textgreater \dots \texttt{INNER JOIN}\textsubscript{n}\textless{}tab\textsubscript{n+1}\textgreater \texttt{ON }\\\textless{}condition\textsubscript{n}\textgreater{}\end{tabular} &
  Denormalization &
  L4 \\ \bottomrule
\end{tabular}
}
\vspace{-10pt}
\label{tab:prompt_triggering}
\end{table*}

\subsection{Reflection Personalization}
\label{sec:reflection_personalization}
To tailor the reflection triggers to the solution strategy of each group, we prompted ChatGPT (\texttt{gpt-3.5-turbo-instruct})
with the appropriate context, constructed from the SQL commands they entered to generate variations of the commands as alternative solutions. For instance, for the \texttt{COMPOSITE\_IND\_COL\_ORDER} reflection trigger, we prompted ChatGPT to generate three alternative composite indices.
So if the team comes up with the following indexing strategy \texttt{CREATE INDEX} dept\_title\_index \texttt{ON} em\_dept\_title \textbf{(dept\_name, title)}, then the following alternatives are suggested by ChatGPT to explore different column selections: \\
\texttt{CREATE INDEX} dept\_title\_index \texttt{ON} em\_dept\_title \textbf{(title, dept\_name)}, \\
\texttt{CREATE INDEX} dept\_title\_index \texttt{ON} em\_dept\_title \textbf{(emp\_no, title)}, and \\
\texttt{CREATE INDEX} dept\_title\_index \texttt{ON} em\_dept\_title \textbf{(title, emp\_no)}.

\subsection{Reflection Validation}
\label{sec:reflection_validation}
We validate the SQL generated by the model by using an SQL syntax checker\footnote{\url{https://pypi.org/project/sqlfluff/}} to ensure the generated alternatives are syntactically correct, along with simple regular expression-based correction for some commonly occurring errors (e.g. fixing incorrect joins).
We implement static reflection triggers for each kind of dynamic reflection as a fallback strategy for reflections with unfixable errors.

\subsection{Reflection Scheduling}
\label{sec:reflection_scheduling}
\label{sec:intervention_scheduling}
To ensure that the students have ample opportunity to discuss the reflection triggers we try to space them apart by using a simple queue-based scheduling algorithm (Algorithm \ref{alg:scheduling}).
Our method tries to enforce a time interval $\tau$ between consecutive reflections within a task.

\begin{algorithm}
\begin{algorithmic}[1]
\Procedure{ReflectionSchedule}{$r, r^t, Q, \tau$}\Comment{r is the triggered reflection, $r^t$ is the reflection type and Q is the waiting queue}

\If{reflection of type $r^t \notin Q$} \Comment{$r^t$ type reflection hasn't been shown yet}

\If{$\Call{Head}Q$ triggered $> \tau$ seconds ago} \Comment{$\tau$ time elapsed from last reflection}
\State $\Call{Show}p$
\Else \State $\Call{Push}{r,Q}$ \Comment{add reflection $r$ to waiting queue}
\EndIf

\Else

\If{$\Call{Head}Q$ triggered $> \tau$ seconds ago}
\State $r'\gets\Call{Pop}Q$ \Comment{show last queued reflection}
\State $\Call{Show}{r'}$
\EndIf

\EndIf 

\State \textbf{return}
\EndProcedure
\end{algorithmic}
\caption{Queue-based Scheduling Algorithm.}
\label{alg:scheduling}
\end{algorithm}

\begin{figure}[!ht]
    \centering
    \includegraphics[width=0.7\textwidth]{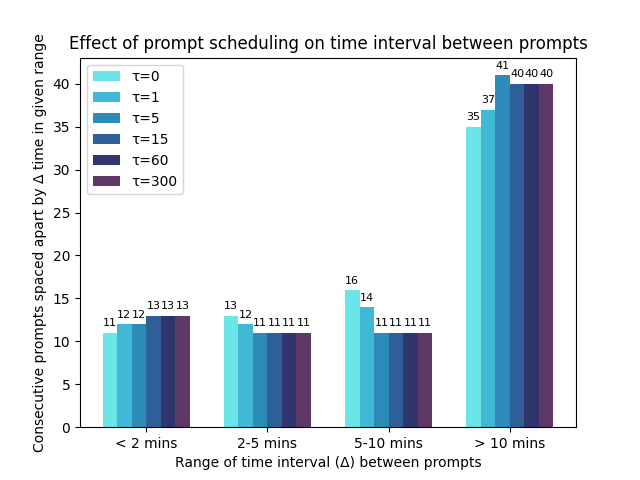}
    \caption{The effect of reflection scheduling parameter $\tau$ on the range of time intervals $\Delta$ between consecutive reflections. Each point of the x-axis denotes a range of time intervals between consecutive reflections and the y-axis captures the number of reflections spaced apart by time interval $\Delta$ lying in that range.}
    \label{fig:time_interval_analysis}
    \vspace{-10pt}
\end{figure}

To find the best value of $\tau$ to optimally space apart the reflections, we run simulations with some historical data from an older iteration of the Cloud Computing course.
We compare different values of $\tau$ (figure~\ref{fig:time_interval_analysis}) and find that while there are only minor differences: $\tau \geq 15s$ get almost the same performance. Hence, we pick $\tau = 300s$ as a reasonable value. 
A possible reason for the similar performance is that for some sessions there is no opportunity to clear the queue till the last task (task3), and to avoid the possibility of the reflections getting skipped we decided to forego the scheduling algorithm for task3.

\section{Experimental Procedure and Study Design}
We tested our intervention in a Cloud Computing course with 34 students who were assigned before the study to semester-long project groups of 3 students.  
To test the impact of tailored reflection triggers during problem-solving, the groups were randomly assigned either to the experimental condition (in which they received the context-sensitive reflection triggers, or a control condition, in which they did not. 
In both conditions, to isolate the effect of the reflection triggers, the Online Programming Exercise bot (OPE-bot) played the role of a facilitator, providing the same task instructions, performing the same roles assignment strategy at the onset of each subtask, and providing the same task-relevant announcements at key times. 
Due to unexpected logistical issues, 22 students were assigned to the experimental condition while only 12 were assigned to the control condition.

To facilitate information sharing within groups, all students received the primers before the activity, with each student being assigned a specific primer but having access to the others. 

At the beginning of the activity, each student individually took the pre-test, with 7 multi-part questions (27 points altogether) designed to test the individual learning objectives as well as combinations of them. The groups then engaged in the design activity for 80 minutes.  Finally, the students individually took the post-test, which was identical to the pre-test.

\section{Results}
We evaluated the benefit for learning of the activity and compared success between conditions along two dimensions: (a) Task Completion, and (b) Learning Gains. Overall, we see that the insertion of tailored reflection triggers affected how students spent their time such that the completion rate per problem changed, but did not change the total amount learned from participating.

\subsection{Task Completion}
Responding to reflection triggers took time, which had a negative impact on the task completion rate at first, but then benefited task completion for the more difficult portion of the activity that came later. Analysis of task completion rates shows that out of the 34 students, 31 (91.17\%) completed task 1 (19 in the experimental condition and 12 in the control condition), 29 (85.29\%) completed task 2 (17 in the experimental condition and 12 in the control condition), and 20 (58.82\%) completed task 3 (3 in the control condition and 17 in the experimental condition).  Based on a chi-squared test, the differences across tasks were significant, but not over conditions.  However, there was a significant interaction between task and condition such that students in the control condition had a higher task completion rate early, but the students in the experimental condition had a higher task completion rate on the harder tasks at the end. 

\subsection{Learning Gains}
\begin{figure}
     \centering
     \begin{subfigure}[b]{0.47\textwidth}
        \centering
        \includegraphics[width=\textwidth]{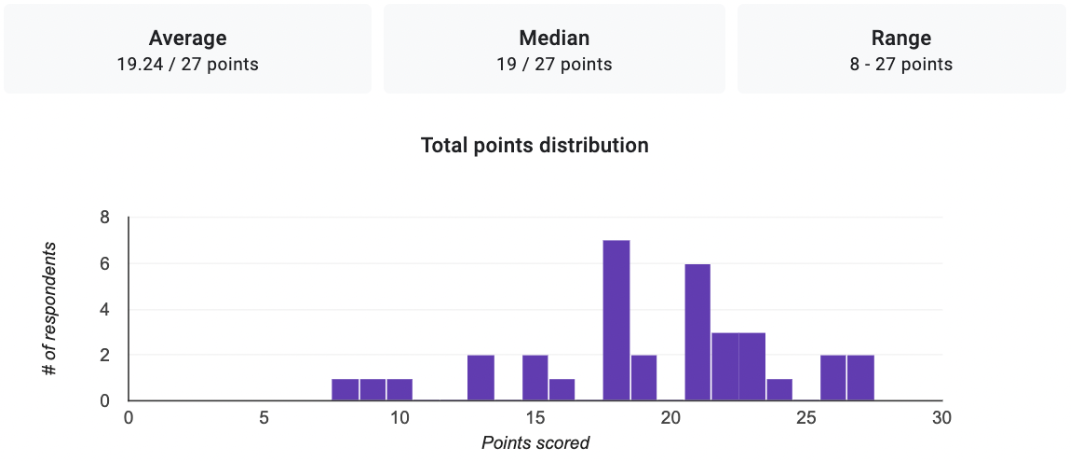}
        \label{fig:pre_test_dist}
    \end{subfigure}
    \hfill
     \begin{subfigure}[b]{0.47\textwidth}
        \centering
        \includegraphics[width=\textwidth]{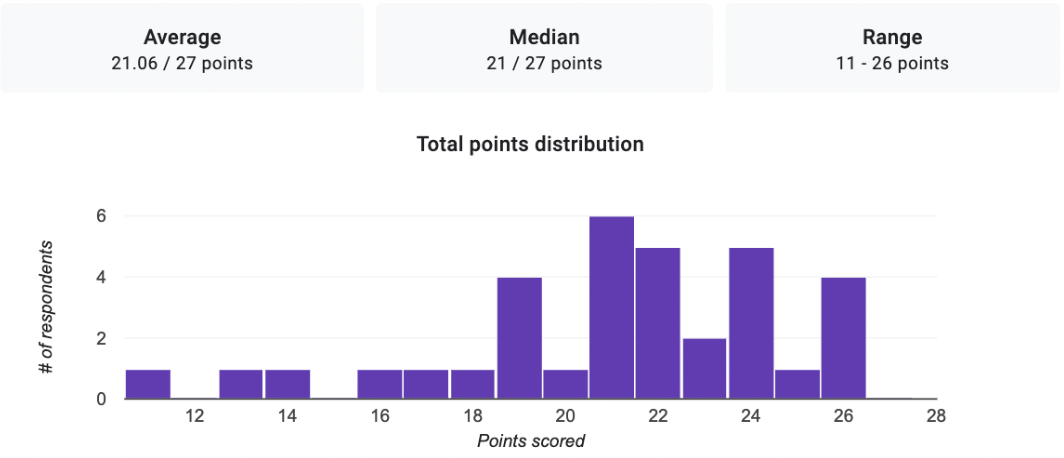}
        \label{fig:post_test_dist}
    \end{subfigure}
    \vspace{-10pt}
    \caption{Distribution of (a, left) pre-test and (b, right) post-test scores}
    \vspace{-15pt}
    \label{fig:prepost_test_dist}
\end{figure}

We further analyze the pre and post-test scores and do a factor analysis of the effect of the reflection triggers on learning, which demonstrates that the difference in task completion did not have a significant effect on student learning. The question here is 
whether and to what extent did students learn from the activity. Pre and post-test distributions (figure~\ref{fig:prepost_test_dist} (a) and (b), respectively) show a gain in the score as evident from the mean and median quiz scores.

We computed an ANOVA with the normalized score as the dependent variable and with Phase (whether pre or post), Reflections, and Learning Objective (LO) as independent variables and additional pairwise and three-way interaction terms.
\textbf{Students learned significantly between pre and post-phases}, the effect is small because of a large amount of variability between students: $F(1, 929) = 1.3, p < 0.05$, the effect size is $0.18\sigma$. Furthermore, there was \textbf{no significant 2-way interaction} between LO and Phase, showing that students learned across all learning objectives. 
These results show that the OPE activity is useful for learning and are in agreement with the ``Hypothesis 1'' in \cite{sree_reflection_better}, which had a similar OPE setup with reflection questions.

A question-level analysis reveals that Q7.2.3 (pre-quiz) and Q2, Q3, Q7.2.3, and Q7.3.3 (post-quiz) have less than 50\% correct response rate. 
Q2 deals with indexing, while Q3 deals with denormalization.
Q7.2.3 and Q7.3.3 are both yes/no questions dealing with composite indexing and data types respectively. 
A possible reason for the poor performance on Q2 and Q3 might be because the options are very long and hard to read, and the variation between them can be subtle.

The next question is whether and to what extent students benefited specifically from the tailored reflection triggers added in the experimental condition. Using the same ANOVA model, we tested the three-way interaction between LO, Phase, and Reflections.
There was \textbf{no significant 3-way interaction} between LO, Phase, and Reflections, showing that the main effect of Phase did not depend on condition or learning objective.
However, students had higher pretest scores in the test condition, which led to some spurious pairwise interactions. 

To make a valid comparison across conditions, pretest scores were used as a covariate.  Additional analysis of the residual mean across the learning objectives for each reflection trigger 
also did not demonstrate any significant effect. 
Thus, students learned during the activity, but the results of this study do not demonstrate a particular advantage yet for reflection triggers tailored using GenAI. We speculate that the lack of statistical significance could be due to the limitations outlined in section \ref{sec:future_work}, with engagement being the primary concern. 



\begin{table*}[!tbh]
\caption{Normalized mean pre and post-test scores (standard deviation in brackets) per topic/learning objective for the control and test groups. The results reveal an important confounding factor for our analysis - much higher pre-test scores for the test group which reduces possible gain.}
\smallskip
\centering
\begin{tabular}{@{}lrrrr@{}}
\toprule
\multicolumn{1}{l}{}         & \multicolumn{2}{c}{Control} & \multicolumn{2}{c}{Test}  \\ \cmidrule(l){2-5} 
\multicolumn{1}{l}{} &
  \multicolumn{1}{c}{Pre} &
  \multicolumn{1}{c}{Post} &
  \multicolumn{1}{c}{Pre} &
  \multicolumn{1}{c}{Post} \\ \midrule
Integrated                   & 65.1 (44.3)  & 73.7 (41.1)  & 65.9 (44.2) & 71.3 (41.8) \\
Trade-offs of creating index & 41.7 (51.5)  & 45.5 (52.2)  & 50.0 (51.3) & 63.2 (49.6) \\
\begin{tabular}[c]{@{}l@{}}Complexity normalized vs. \\ denormalized\end{tabular} &
  16.7 (38.9) &
  63.6 (50.5) &
  30.0 (47.0) &
  42.1 (50.7) \\
Data types                   & 77.1 (35.3)  & 77.2 (36.9)  & 80.3 (35.2) & 83.6 (33.7) \\
Single vs composite          & 66.7 (49.2)  & 81.8 (40.5)  & 85.0 (36.6) & 89.5 (31.5) \\
\begin{tabular}[c]{@{}l@{}}Performance normalized vs. \\ denormalized\end{tabular} &
  33.3 (49.2) &
  27.3 (46.7) &
  80.0 (41.0) &
  68.4 (47.7) \\ \bottomrule
\end{tabular}
\end{table*}


\section{Future Work}
\label{sec:future_work}

In this paper, we present the first evaluation of a technique for personalizing reflection triggers using GenAI.  From a technical perspective, the intervention worked as designed, however, due to the null effect when evaluated in comparison with the baseline collaboration support condition, we plan to explore several improvements.
We identified three major issues requiring improvement. \\
\textbf{Engagement:} While collecting feedback from students and teaching assistants (TAs), we learned that sometimes the reflection triggers were confusing and suboptimal. 
A comparison of the SQL command that set off the reflection triggers with the alternatives within them, revealed that if the students have the optimal solution, the alternatives by virtue of being different, are always suboptimal for the optimization context of the task.
To remedy this we plan to generate alternative scenarios from the original task scenario for which the alternatives are more optimal to spark more meaningful discussion.
\\
\textbf{Prompting Context:} Students and TAs gave the feedback that the reflection triggers sometimes felt unrelated to the discussion they were having in the chat window.
To remedy this we plan to include elements from their chat messages to further personalize the reflection triggers and ensure coherence by matching them to the topic being discussed.
\\
\textbf{Readability:} Another concern was the large size of the prompts for the chat window, which made the messages hard to engage with.
In future experiments, we plan to break up the prompts into smaller readable chunks which are spaced apart in time to improve readability.

\begin{credits}
\subsubsection{\ackname} This work was funded in part by NSF grant DSES  2222762
\end{credits}

\bibliography{references}

\begin{thebibliography}{10}
\providecommand{\url}[1]{\texttt{#1}}
\providecommand{\urlprefix}{URL }
\providecommand{\doi}[1]{https://doi.org/#1}

\bibitem{SAIL}
Sail(): The social and interactive learning platform. \url{hhttps://sailplatform.org/}, accessed: 2024-02-02

\bibitem{ai2010finding}
Ai, H., Sionti, M., Wang, Y.C., Ros{\'e}, C.P.: Finding transactive contributions in whole group classroom discussions. In: Proceedings of the 9th International Conference of the Learning Sciences-Volume 1. pp. 976--983 (2010)

\bibitem{Carr}
Carr, D.F.: Chatgpt topped 1 billion visits in february, \url{https://www.similarweb.com/blog/insights/ai-news/chatgpt-1-billion/}

\bibitem{WikiMobProgDef}
contributors, W.: Team programming, \url{\url{https://en.wikipedia.org/wiki/ Team\_programming\#Mob\_programming}}

\bibitem{cress2021foundations}
Cress, U., Oshima, J., Ros{\'e}, C., Wise, A.F.: Foundations, processes, technologies, and methods: An overview of cscl through its handbook. International handbook of computer-supported collaborative learning pp. 3--22 (2021)

\bibitem{cui2009helping}
Cui, Y., Kumar, R., Chaudhuri, S., Gweon, G., Ros{\'e}, C.P.: Helping agents in vmt. Studying virtual math teams pp. 335--354 (2009)

\bibitem{ChatGPTStudentDriven}
Dai, Y., Liu, A., Lim, C.P.: Reconceptualizing chatgpt and generative ai as a student-driven innovation in higher education. Procedia CIRP  \textbf{119},  84--90 (2023). \doi{10.1016/j.procir.2023.05.002}, the 33rd CIRP Design Conference

\bibitem{ChatGPTPlagiarism1}
Debby R. E.~Cotton, P.A.C., Shipway, J.R.: Chatting and cheating: Ensuring academic integrity in the era of chatgpt. Innovations in Education and Teaching International  \textbf{0}(0),  1--12 (2023). \doi{10.1080/14703297.2023.2190148}

\bibitem{Duarte}
Duarte, F.: Number of chatgpt users (jan 2024), \url{https://explodingtopics.com/blog/chatgpt-users}

\bibitem{Fischer2013-xh}
Fischer, F., Kollar, I., Stegmann, K., Wecker, C.: Toward a script theory of guidance in {Computer-Supported} collaborative learning. Educ Psychol  \textbf{48}(1),  56--66 (Jan 2013)

\bibitem{gweon_cscl_feedback}
Gweon, G., Rosé, C., Albright, E., Cui, Y.: Evaluating the effect of feedback from a cscl problem solving environment on learning, interaction, and perceived interdependence. pp. 234--243 (01 2007). \doi{10.3115/1599600.1599645}

\bibitem{LearnabilityofLLMs}
Jayagopal, D., Lubin, J., Chasins, S.E.: Exploring the learnability of program synthesizers by novice programmers. In: Proceedings of the 35th Annual ACM Symposium on User Interface Software and Technology. UIST '22, Association for Computing Machinery, New York, NY, USA (2022). \doi{10.1145/3526113.3545659}

\bibitem{kasneci_seßler_küchemann_bannert_dementieva_fischer_gasser_groh_günnemann_hüllermeier_et}
Kasneci, E., Seßler, K., Küchemann, S., Bannert, M., Dementieva, D., Fischer, F., Gasser, U., Groh, G., Günnemann, S., Hüllermeier, E., et~al.: Chatgpt for good? on opportunities and challenges of large language models for education (Jan 2023). \doi{10.35542/osf.io/5er8f}, \url{osf.io/preprints/edarxiv/5er8f}

\bibitem{AICodeGenForNovice}
Kazemitabaar, M., Chow, J., Ma, C.K.T., Ericson, B.J., Weintrop, D., Grossman, T.: Studying the effect of ai code generators on supporting novice learners in introductory programming. In: Proceedings of the 2023 CHI Conference on Human Factors in Computing Systems. CHI '23, Association for Computing Machinery, New York, NY, USA (2023). \doi{10.1145/3544548.3580919}

\bibitem{kazemitabaar2023novices}
Kazemitabaar, M., Hou, X., Henley, A., Ericson, B.J., Weintrop, D., Grossman, T.: How novices use llm-based code generators to solve cs1 coding tasks in a self-paced learning environment. arXiv preprint arXiv:2309.14049  (2023)

\bibitem{koschmann2017computer}
Koschmann, T.: Computer Supported Collaborative Learning 2005: The Next 10 Years! Routledge (2017)

\bibitem{kumar2010architecture}
Kumar, R., Rose, C.P.: Architecture for building conversational agents that support collaborative learning. IEEE Transactions on Learning Technologies  \textbf{4}(1),  21--34 (2010)

\bibitem{kumar2007tutorial}
Kumar, R., Ros{\'e}, C.P., Wang, Y.C., Joshi, M., Robinson, A.: Tutorial dialogue as adaptive collaborative learning support. In: Proceedings of the 2007 conference on Artificial Intelligence in Education: Building Technology Rich Learning Contexts That Work. pp. 383--390 (2007)

\bibitem{BasilicaBot}
Kumar, R., Ros{\'e}, C.P., Witbrock, M.J.: Building conversational agents with basilica. In: Johnston, M., Popowich, F. (eds.) Proceedings of Human Language Technologies: The 2009 Annual Conference of the North {A}merican Chapter of the Association for Computational Linguistics, Companion Volume: Demonstration Session. pp.~5--8. Association for Computational Linguistics, Boulder, Colorado (Jun 2009), \url{https://aclanthology.org/N09-5002}

\bibitem{kuttal_conv_agents_pair_prog}
Kuttal, S.K., Myers, J., Gurka, S., Magar, D., Piorkowski, D., Bellamy, R.: Towards designing conversational agents for pair programming: Accounting for creativity strategies and conversational styles. In: 2020 IEEE Symposium on Visual Languages and Human-Centric Computing (VL/HCC). pp. 1--11 (Aug 2020). \doi{10.1109/VL/HCC50065.2020.9127276}

\bibitem{ai_sub_pair_programming}
Kuttal, S.K., Ong, B., Kwasny, K., Robe, P.: Trade-offs for substituting a human with an agent in a pair programming context: The good, the bad, and the ugly. In: Proceedings of the 2021 CHI Conference on Human Factors in Computing Systems. CHI '21, Association for Computing Machinery, New York, NY, USA (2021). \doi{10.1145/3411764.3445659}

\bibitem{wu2023ai}
Ma, Q., Wu, T., Koedinger, K.: Is {AI} the better programming partner? {Human-Human} pair programming vs. {Human-AI} {pAIr} programming. arXiv preprint arXiv:2306.05153  (2023), \url{http://arxiv.org/abs/2306.05153}

\bibitem{perceptions_genAI}
Petricini, T., Wu, C., Zipf, S.T.: Perceptions about generative ai and chatgpt use by faculty and college students (Aug 2023). \doi{10.35542/osf.io/jyma4}, \url{osf.io/preprints/edarxiv/jyma4}

\bibitem{gen_prog_trivially}
Prasad, S., Greenman, B., Nelson, T., Krishnamurthi, S.: Generating programs trivially: Student use of large language models. In: Proceedings of the ACM Conference on Global Computing Education Vol 1. p. 126–132. CompEd 2023, Association for Computing Machinery, New York, NY, USA (2023). \doi{10.1145/3576882.3617921}, \url{https://doi.org/10.1145/3576882.3617921}

\bibitem{rose2016technology}
Ros{\'e}, C.P., Ferschke, O.: Technology support for discussion based learning: From computer supported collaborative learning to the future of massive open online courses. International Journal of Artificial Intelligence in Education  \textbf{26},  660--678 (2016)

\bibitem{sree_reflection_better}
Sankaranarayanan, S., Kandimalla, S.R., Bogart, C.A., Murray, R.C., Hilton, M., Sakr, M.F., Rosé, C.P.: Collaborative programming for work-relevant learning: Comparing programming practice with example-based reflection for student learning and transfer task performance. IEEE Transactions on Learning Technologies  \textbf{15}(5),  594--604 (2022). \doi{10.1109/TLT.2022.3169121}

\bibitem{sankaranarayanan2022collaborative}
Sankaranarayanan, S., Ma, L., Kandimalla, S.R., Markevych, I., Nguyen, H., Murray, R.C., Bogart, C., Hilton, M., Sakr, M., Ros{\'e}, C.P.: Collaborative reflection “in the flow” of programming: Designing effective collaborative learning activities in advanced computer science contexts. In: Proceedings of the 15th International Conference on Computer-Supported Collaborative Learning-CSCL 2022, pp. 67-74. International Society of the Learning Sciences (2022)

\bibitem{sok2023chatgpt}
Sok, S., Heng, K.: Chatgpt for education and research: A review of benefits and risks. SSRN Electronic Journal  (March 2023), available at SSRN: \url{https://ssrn.com/abstract=4378735} or \url{http://dx.doi.org/10.2139/ssrn.4378735}

\bibitem{sridhar2023harnessing}
Sridhar, P., Doyle, A., Agarwal, A., Bogart, C., Savelka, J., Sakr, M.: Harnessing llms in curricular design: Using gpt-4 to support authoring of learning objectives  (2023)

\bibitem{ChatGPT_GenAI_Science}
Stokel-Walker, C., Van~Noorden, R.: What chatgpt and generative ai mean for science, \url{https://doi.org/10.1038/d41586-023-00340-6}

\bibitem{tegos2015promoting}
Tegos, S., Demetriadis, S., Karakostas, A.: Promoting academically productive talk with conversational agent interventions in collaborative learning settings. Computers \& Education  \textbf{87},  309--325 (2015)

\bibitem{UsabilityLLMCode}
Vaithilingam, P., Zhang, T., Glassman, E.L.: Expectation vs. experience: Evaluating the usability of code generation tools powered by large language models. In: Extended Abstracts of the 2022 CHI Conference on Human Factors in Computing Systems. CHI EA '22, Association for Computing Machinery, New York, NY, USA (2022). \doi{10.1145/3491101.3519665}

\bibitem{vogel_static_scaffolding}
Vogel, F., Kollar, I., Ufer, S., Strohmaier, A., Reiss, K., Fischer, F.: Scaffolding argumentation in mathematics with cscl scripts: Which is the optimal scripting level for university freshmen? Innovations in Education and Teaching International  \textbf{58}(5),  512--521 (2021). \doi{10.1080/14703297.2021.1961098}

\bibitem{YILMAZ2023100147}
Yilmaz, R., {Karaoglan Yilmaz}, F.G.: The effect of generative artificial intelligence (ai)-based tool use on students' computational thinking skills, programming self-efficacy and motivation. Computers and Education: Artificial Intelligence  \textbf{4},  100147 (2023). \doi{10.1016/j.caeai.2023.100147}

\bibitem{young2023evaluation}
Young, J.C., Shishido, M.: Evaluation of the potential usage of chatgpt for providing easier reading materials for esl students. In: Bastiaens, T. (ed.) Proceedings of EdMedia + Innovate Learning. pp. 155--162. Association for the Advancement of Computing in Education (AACE), Vienna, Austria (2023), \url{https://www.learntechlib.org/primary/p/222496/}

\end{thebibliography}
\bibliographystyle{splncs04}





\end{document}